\documentclass[runningheads]{llncs}
\usepackage{graphicx}

\usepackage{hyperref}

\begin{document}

\title{Investigating the generative dynamics of energy-based
neural networks\thanks{Supported by grant RF-2019-12369300 from the Italian Ministry of Health.}}

\titlerunning{Generative dynamics of energy-based neural networks}

\author{Lorenzo Tausani\inst{1,2}\and
Alberto Testolin\inst{1,2}\orcidID{0000-0001-7062-4861} \and
Marco Zorzi\inst{2,3}\orcidID{0000-0002-4651-6390}}

\authorrunning{Tausani et al.}
\institute{
Department of Mathematics, University of Padova, 35141 Padua, Italy
\and
Department of General Psychology and Padova Neuroscience Center, University of Padova, 35141 Padua, Italy \and
IRCCS San Camillo Hospital, 30126 Venice Lido, Italy
\email{\{alberto.testolin,marco.zorzi\}@unipd.it}}

\maketitle

\begin{abstract}
Generative neural networks can produce data samples according to the statistical properties of their training distribution. This feature can be used to test modern computational neuroscience hypotheses suggesting that spontaneous brain activity is partially supported by top-down generative processing.
A widely studied class of generative models is that of Restricted Boltzmann Machines (RBMs), which can be used as building blocks for unsupervised deep learning architectures. In this work, we systematically explore the generative dynamics of RBMs, characterizing the number of states visited during top-down sampling and investigating whether the heterogeneity of visited attractors could be increased by starting the generation process from biased hidden states.
By considering an RBM trained on a classic dataset of handwritten digits, we show that the capacity to produce diverse data prototypes can be increased by initiating top-down sampling from chimera states, which encode high-level visual features of multiple digits. We also found that the model is not capable of transitioning between all possible digit states within a single generation trajectory, suggesting that the top-down dynamics is heavily constrained by the shape of the energy function.

\end{abstract}

\section{Introduction}

One frontier of modern neuroscience is understanding the so-called \textit{spontaneous brain activity}, which arises when the brain is not engaged in any specific task \cite{Mitra et al. 2018}. This intrinsic activity accounts for most of brain energy consumption \cite{Mitra Raichle 2016}, and has been studied using electrophysiological recordings \cite{Pan et al. 2011}, electroencephalography \cite{Tortella-Feliu et al. 2014} and functional magnetic resonance imaging \cite{Leuthardt et al. 2015}.

A recently proposed computational framework \cite{Pezzulo et al. 2021} suggests that spontaneous activity could be interpreted as top-down computations that occur in \textit{generative models}, whose goal is to estimate the latent factors underlying the observed data distribution \cite{Parr and Friston 2018}. This framework entails a strong connection between spontaneous and task-related brain activity: when performing a task, the generative model would focus on maximizing accuracy in the task of interest, while during rest the model would reproduce task-related activation patterns and use them for the computation of generic spatiotemporal priors that summarize a large variety of task representations with a low dimensionality \cite{Pezzulo et al. 2021}. This is in agreement with modeling work suggesting that the brain at rest is in a state of maximum metastability \cite{Deco et al. 2017}, where brain regions are organized into quasi-syncronous activity, interrupted by periods of segregation, without getting caught in attractor states \cite{Tognoli Kelso 2014}.

Deep learning models are increasingly used to simulate the activity of biological brains and explore the principles of neural computation \cite{Richards et al. 2019,De Schutter 2018}.
For example, deep networks have been used to reproduce some functional properties of cortical processing, particularly in the visual system \cite{Yamins DiCarlo 2016}, as well as to simulate a variety of cognitive functions (e.g., \cite{Stoianov Zorzi 2012,Zorzi et al. 2013,Testolin et al. 2017}) and their progressive development \cite{Testolin et al. 2020,Zambra et al. 2022}.
However, it is not well understood whether existing deep learning architectures could capture key signatures of spontaneous brain activity. 

Here we propose to investigate the (spontaneous) generative dynamics of a well-known class of generative models called Restricted Boltzmann Machines (RBMs), which are a particular type of energy-based neural networks rooted in statistical physics \cite{Ackley et al. 1985}. 
The RBM is an undirected graphical model formed by two layers of symmetrically connected units. Visible units encode the data (e.g., pixels of an image), whereas hidden units discover latent features through unsupervised generative learning \cite{Ackley et al. 1985}. In RBMs, sampling from the hidden states leads to generating visible states that correspond to trained patterns, but these configurations represent local energy minima (i.e., attractors) that are difficult to escape. Indeed, large energy barriers need to be crossed to go from one (stable) visible state to another, which makes these transitions very difficult \cite{Roussel et al. 2021}.

Our approach aims at finding constrained initializations of hidden states that could induce the network into metastable sample generation, thus simulating the dynamics of spontaneous activity in the brain. We quantify this as the number of digit states explored in a generation round, identified by a trained neural network classifier, avoiding to get caught in attractor states. 
In the first set of simulations, we exploit the method described in \cite{Zorzi et al. 2013} to sample visual patterns starting from hidden states derived by inverting a classifier trained to map internal representations into one-hot encoded labels. Next, we describe two variations of the original method that combine features of different digits to produce ``biased'' hidden states away from attractor basins, which should be capable of exploring more states during the generation process. Our results indicate that such biased states indeed increase state exploration compared to classical label biasing with digit labels. However, no hidden state is capable of inducing the exploration of all digits in a single generation round, suggesting that the RBM in its classic version is not capable of mimicking the continuous and heterogeneous state exploration demonstrated by biological brains.

\section{Materials and Methods}

\subsection{Dataset}
Our simulations are based on the classic MNIST dataset \cite{LeCun et al. 1998}, which contains images of 28x28 pixels representing handwritten digits from 0 to 9, encoded in 8-bit grayscale (values from 0 to 255, normalized between 0 and 1). It encompasses a training set of 60000 examples and a testing set of 10000 examples. Although this is a medium-sized dataset with a limited number of classes, it allows us to more clearly characterize the generative dynamics by measuring the number of different states visited during top-down sampling.

\subsection{Restricted Boltzmann machines}
Boltzmann machines are energy models composed of two different kinds of units: \textit{visible units}, which are used to provide input data (e.g. pixels of an image) and \textit{hidden units}, which are used to extract latent features by discovering higher-order interactions between visible units \cite{Ackley et al. 1985}. In RBMs there are no hidden-to-hidden and visible-to-visible connections: the only connections are between the visible and hidden units, which can be considered as two separate layers of a bipartite, fully-connected graph \cite{Goodfellow et al. 2016}.
Neurons in a Boltzmann machine are conceptualized as stochastic units, whose activity is the result of a Bernoullian sampling with activation probability $P\left(\sigma_i=1\right)$ defined as follows: 

\begin{equation} \label{eq:Stoc_unit_activation}
P\left(\sigma_i=1\right)=\frac{1}{1+e^{-\Delta E_i / T}}
\end{equation}

\noindent where $\Delta E_i$ is the difference in the energy of the system caused by the change in the state of the unit $i$, and $T$ is the temperature parameter that acts as a noise factor. 
Given a set of training data $\mathcal{D}=\left\{{x}^{(i)}\right\}_{i=1}^n$ the parameters $\theta$ of an RBM (that is, the weights connecting the units and the biases) are updated by maximizing the likelihood $p(\mathcal{D} | \theta)$, where $p(\mathcal{D} | \theta)$ is the Boltzmann distribution with temperature $T=1$. Training is performed by gradient ascent, usually adopting the contrastive divergence training algorithm, which exploits Monte Carlo Markov chain methods to estimate the gradient update \cite{Hinton 2002}.

\subsubsection{Model architecture and training details}
In our study, we used an RBM with 784 visible units (that is, equal to the vectorization of single MNIST examples (28x28 = 784)) and 1000 hidden units. The RBM was trained with 1 step contrastive divergence and learning rate $\eta = 0.1$ for both weights and biases (hidden and visible). The parameter update also included a momentum term $\gamma$ to speed up the training. Following standard practice \cite{Testolin et al. 2013} $\gamma$ was equal to 0.5 in the first 5 training epochs and 0.9 in successive iterations. Furthermore, the parameter update was decreased by the value of the parameter of interest in the previous training iteration multiplied by a decay factor equal to 0.0002. Both hidden and visible biases were initialized equal to 0, while connection weights were initialized with random numbers sampled from a zero-mean normal distribution with standard deviation equal to 0.1. The model was trained for 100 epochs following a batch-wise approach, with batch size = 125. Learning was monitored using a root mean square error loss function. 

\subsubsection{Top-down sampling from RBM}
Data generation was performed at the end of the RBM training phase. To generate smoother images, during top-down sampling visible units were not binarized, thus assuming continuous values between 0 and 1. Hidden units were instead binarized through Bernoulli sampling. Data patterns were generated following the \textit{label biasing} procedure described in \cite{Zorzi et al. 2013}, where examples are generated top-down from a hidden state vector $H_{\mathrm{Label\ biasing}}$ obtained through the inversion of a linear classifier trained to classify the digit class from its hidden representation.

\begin{figure}
\includegraphics[width=\textwidth]{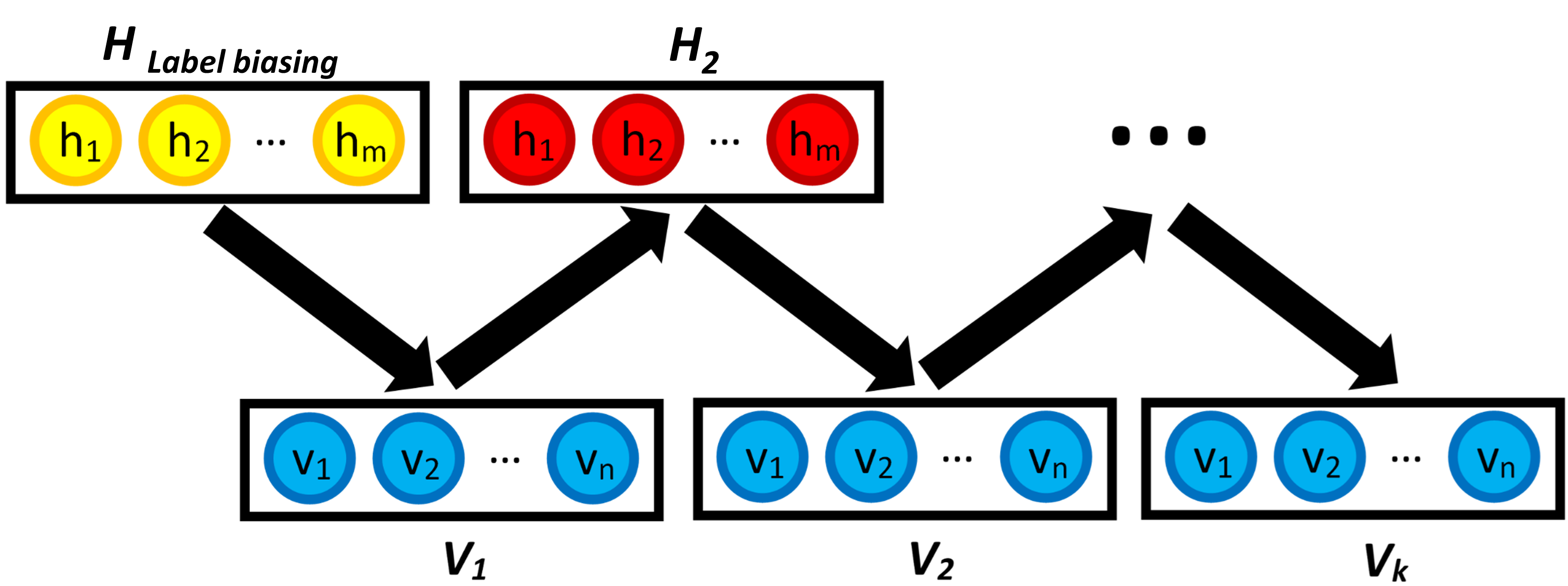}
\caption{Illustration of the label biasing generation procedure. A hidden state vector $H_{\mathrm{Label\ biasing}}$ is obtained using the linear projection method \cite{Zorzi et al. 2013}. Then from $H_{\mathrm{Label\ biasing}}$ a visible vector $V_1$ is generated. The process is repeated $k$ times, where $k$ is the desired number of generation steps.}
\label{Label biasing paradigm}
\end{figure}

A \textit{generation step} is defined as a single generation of a visible state (generated sample) from a hidden state. The generated sample is then used to instantiate the hidden state of the next generation step.
In the first generation step, the activation of the visible layer $A_V$ is computed as the matrix multiplication between $H_{\mathrm{Label\ biasing}}$ and the transposed weight matrix W of the RBM model. The result of the operation is added to the visible bias $b_V$:
\begin{equation}\label{eq:visible_act_label_biasing}
A_V= (H_{\mathrm{Label\ biasing}}\cdot W^{T}) + b_V
\end{equation}

The first visible state $V_1$ is computed as the output of a sigmoid activation function taking as input $A_V$  divided by the temperature $T$:
\begin{equation} \label{eq:visible probability}
V_1=\sigma(\frac{A_V}{T})
\end{equation}  

In the following generation steps, the hidden state $H_s$ is computed as follows: 
\begin{equation}\label{eq:p_hidden}
H_s \sim Bernoulli \left(p=\sigma(\frac{V_{s-1} \cdot W + b_H}{T})\right)
\end{equation} 

\noindent where $V_{s-1}$ is the visible state of the previous reconstruction step and $b_H$ is the hidden bias. The consequent visible states are computed following the same procedure described for step 1 (Fig.~\ref{Label biasing paradigm}).

\subsection{Digit classifier}
In order to establish whether top-down generation resulted in well-formed image patterns over visible units, we trained a classifier to identify digit classes taking as input the patterns generated by the RBM. We used a VGG-16 classifier, which is a convolutional architecture widely used in image classification \cite{Simonyan 2014}. The model was adapted from \footnotemark[1] \footnotetext[1]{\href{https://colab.research.google.com/drive/1IN0HD7-ljlPFtsbstfxLSKWvg2y2ndmO?usp=sharing}{https://colab.research.google.com/drive/1IN0HD7-ljlPFtsbstfxLSKWvg2y2ndmO?usp=sharing}} and was made up of 4 VGG block units, followed by 3 fully connected layers and a final softmax layer. Unlike \footnotemark[1], the final fully connected layer outputted a vector of 11 entries (i.e., the number of MNIST classes plus one special class representing non-digit samples), which was then processed by a softmax layer. Softmax output was used to classify the example and estimate the uncertainty of the network in the classification, which was measured by calculating the entropy of the softmax output. 

The classifier was trained on the MNIST dataset, with grayscale images resized to 32x32 pixels. The training set was made up of 113400 examples: 54000 were extracted from the MNIST training set, while the remaining 59400 represented non-digit examples. This was done to exclude random classifications when the network was exposed to unrecognizable digits, which is a situation that often occurs during spontaneous top-down sampling in energy-based models. Among these non-digit examples, 5400 were composed of scrambled digit images, while the remaining 54000 were training set examples with a random number of adjacent active pixels (i.e. intensity $> 0$) masked. The choice of this method for producing non-digits was motivated by empirical observation of cases in which the RBM generation produced objects that could not be identified as digits by a human observer. Learning was monitored through a validation set made up of the remaining 6000 examples of the MNIST training set. Testing was done on the 10000 images of the MNIST test set. The model was trained using minibatches of size 64, with stochastic gradient descent and learning rate $\eta = 0.01$ with cross-entropy loss. The model was trained for 20 epochs, selecting the model resulting in the highest validation accuracy ($99,3\%$).

\subsection{Generativity metrics}
In order to measure the diversity and stability of the generative dynamics of the model, we implemented several metrics to characterize changes in visual and hidden activation during top-down sampling. The idea is that the model should develop attractor states in correspondence to digit representations, which are then dynamically visited during spontaneous generation of sensory patterns.

For each generation step, the classifier evaluated the class (i.e. digit from 0 to 9 and non-digit case) of the sample produced. The \textit{number of states visited} was defined as the number of different digits visited during the generation process, without including the non-digit state. Multiple visits to the same state (i.e., same digit recognized by the classifier) during a single generation trajectory were counted as 1. A related metric was the number of generation steps (\textit{state time} in short) in which the sample remained in each digit state, including the non-digit state. This index measures the stability of each attractor state. Finally, we measured the \textit{number of transitions} occurring during the generation process. A transition was defined as the change in classification of a sample from one state to another (transitions to the non-digit state were not included in this quantification). Transitions between states, including the non-digit state, were also used to estimate a \textit{transition matrix} of the entire generation procedure (i.e., taking into account all samples and all generation steps). The aim of the transition matrix was to estimate the probability during the generation process to transition from one digit (or non-digit) state to another. The transition matrix was estimated by counting all transitions from one state to another, normalized by the total number of transitions from that particular state.

For each label biasing vector used, 100 samples were generated. For each sample, a generation period of 100 generation steps was performed.
Measures are reported together with standard error of the mean.

\section{Results}

\begin{figure}
     \includegraphics[width=\linewidth]{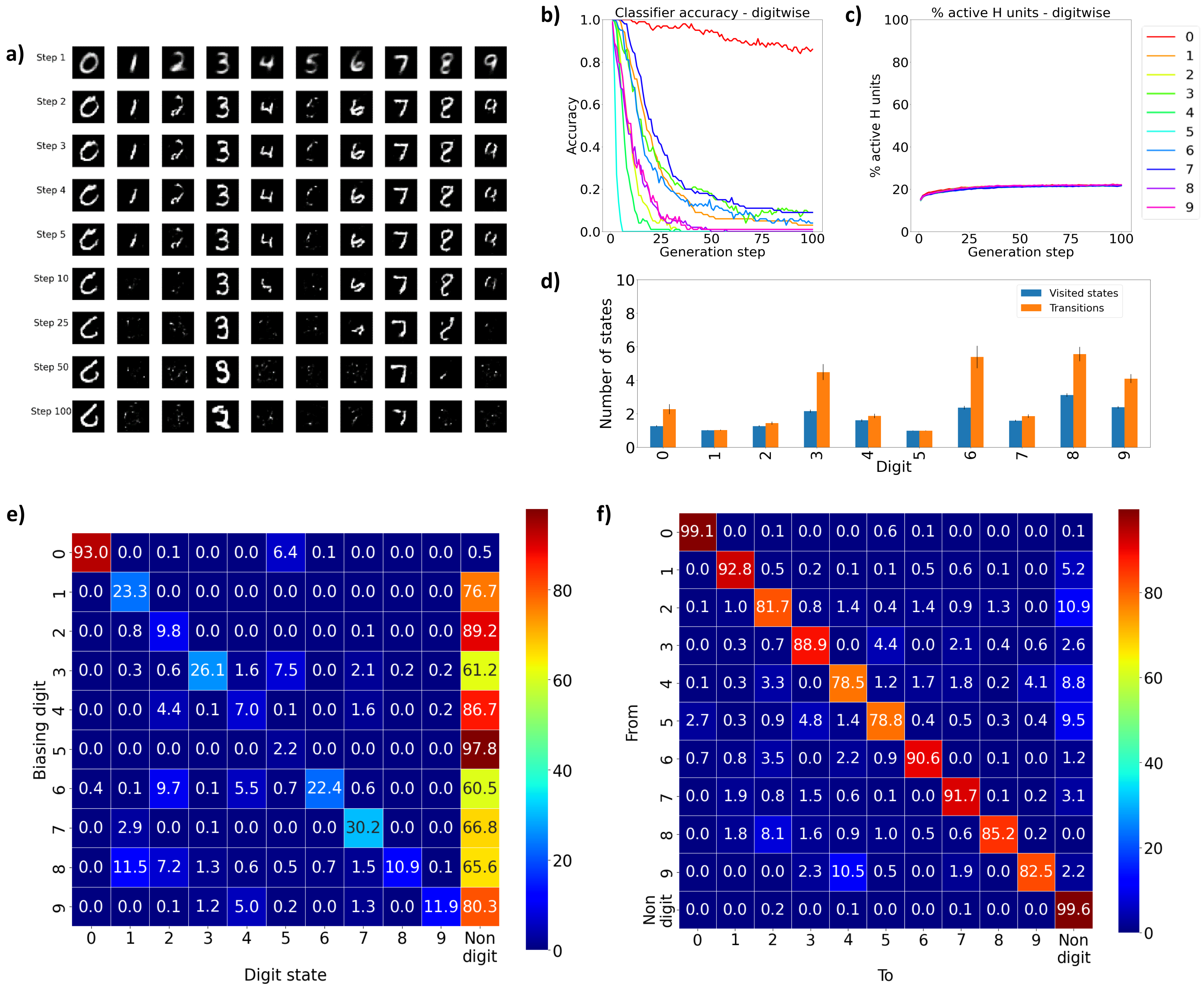}
     \caption{Characterization of sample generation. \textbf{a)} Example of generations, one per digit. Each column represents a single generation in particular generation steps (rows). Accuracy of the classifier (\textbf{b)}), and  average percentage of active hidden units (i.e. $h=1$, \textbf{c)}) as a function of the generation step. Each color represents a different digit. \textbf{d)} Average number of visited states and states transitions per generation period for different label biasing digits (on the $x$ axis). \textbf{e)} Average state time per each digit state (columns) for different label biasing digits (rows). \textbf{f)} Transition matrix estimated from all generated data. Each entry represents the probability of transition from one state (rows) to another (columns).  
     }\label{risultati1}
\end{figure}

The classifier accuracy decreased as a function of the generation step for all digits (average classifier accuracy - step 100: $11.2\%$, Fig.~\ref{risultati1}b), except for the digit zero, which only saw a moderate decrease (classifier accuracy (digit: 0) - step 100: $86.0\%$). This indicated that the samples were significantly distorted during the generation period, inducing more errors in the classifier (examples of sample generation from each digit are shown in Fig.~\ref{risultati1}a). In accordance with this, the average classification entropy increased during the generation period, showing a high anticorrelation with the classifier accuracy ($\rho = -0.999$). Interestingly, all digits showed a similar percentage of active units in the hidden layer throughout the generation process, keeping active only $14-22 \%$ of the units (average percentage of active hidden units - step 1: $14.964 \pm 0.079 \%$, average percentage of active units - step $100$ : $21.892 \pm 0.054 \%$, n = 10 digits, Fig.~\ref{risultati1}c), which is in line with previous results suggesting the emergence of sparse coding in RBM models \cite{Testolin et al. 2017 n2}.

On average, in each generation period $1.779 \pm 0.211$ states were visited, with $2.903 \pm 0.538$ transitions between states (Fig.~\ref{risultati1}d, n = 1000). The transition matrix shows that most transitions occur within the same class of digits (average probability of transition within the same digit: $0.870 \pm 0.021$, $n = 10$, Fig.~\ref{risultati1}f), while the probabilities for a digit state to transition to another digit state are low, almost never exceeding $0.01$ ($0.012 \pm 0.002$, n = 110). This, combined with the small number of transitions per generation period, suggests that state transitions are sharp and that ``bouncing between two states'' events are very rare if not present.
Non-digits transition almost invariably to themselves: in other words, when a sample transitions to a non-digit, it hardly ever gets out of it in the following generation steps. The consequences of this attractor-like behavior of non-digits states is that all digits except 0 spend the majority of the generation period as non-digits (average non-digit state time between digits (0 excluded): $76.099 \pm 4.213$, n=9 digits, Fig.~\ref{risultati1}e). 

A limitation of the ``single digit'' label biasing approach described in the previous paragraph is that it does not allow to explore heterogeneous sensory states, as highlighted by the small number of digit states visited on average in a single generation period (Fig.~\ref{risultati1}d). This might be due to the fact that label biasing forces the RBM to start the generation from a hidden state close to an attractor basin corresponding to the prototype of the selected digit, thus limiting the exploration of other states during top-down sampling. A way to overcome this issue could be to bias the network toward \textit{chimera states}, for example by starting the generation from a hidden state mixing different digit representations. The hypothesis is that this could increase state exploration by decreasing the probability of stranding the generation process in a specific attractor.

\begin{figure}
     \includegraphics[width=\linewidth]{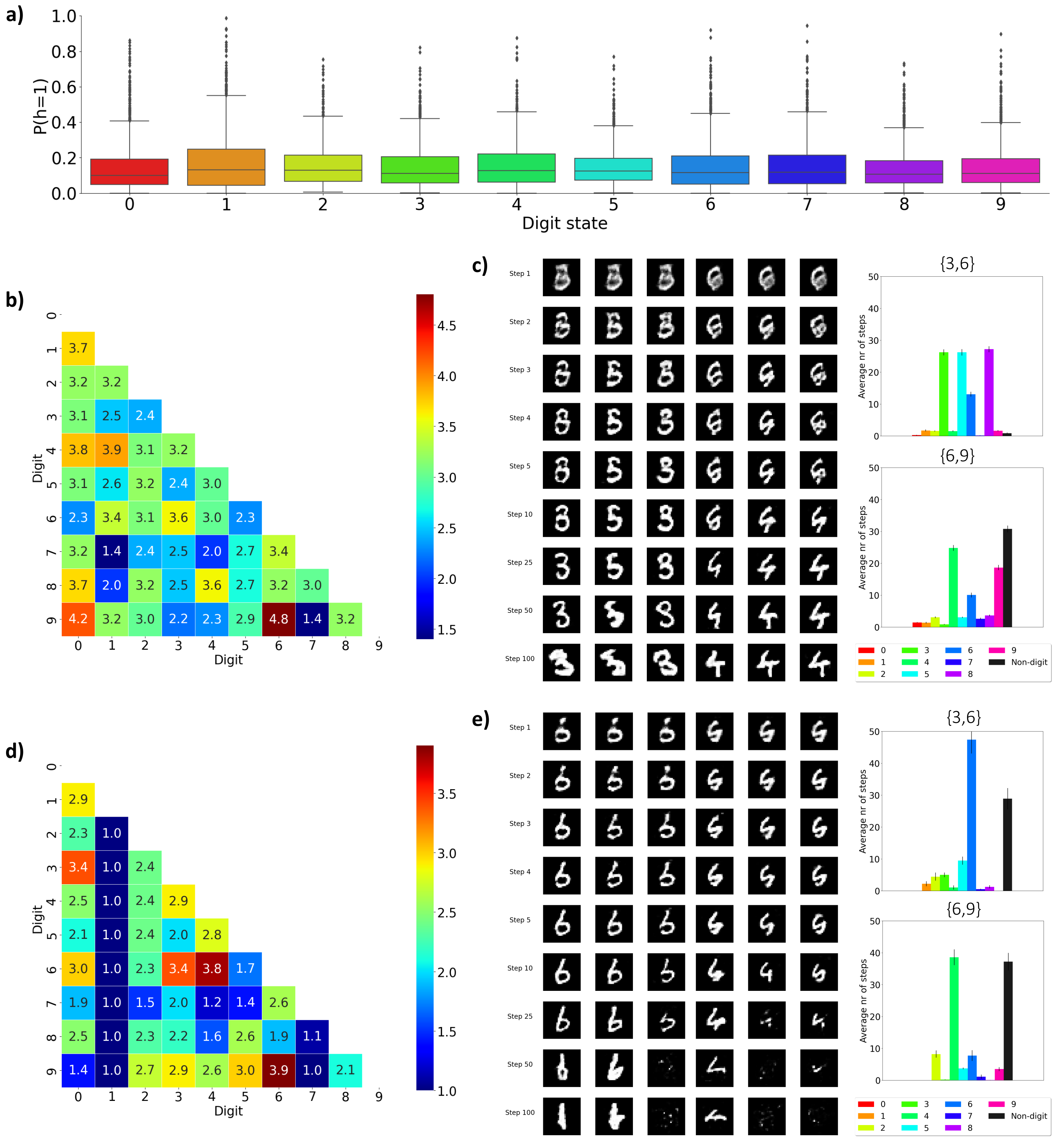}
     \caption{Characterization of generation using chimera states. \textbf{a)} Distribution of activation probability of hidden units ($P(h=1)$) of label biasing vectors of each digit. \textbf{b)} Average number of visited states in a generation period (i.e. 100 generation steps) for each chimera state of two digits using the intersection method (n=100 samples). \textbf{c)} Example generation periods with two example intersection method chimera states (i.e. $\{3,6\}$ (columns 1 to 3) and $\{6,9\}$ (columns 4 to 6)). The average digit state times are shown in the bar plot on the right. \textbf{d)} Average number of visited states in a generation period (i.e. 100 generation steps) for each chimera state of two digits using the double label biasing method (n=100 samples). \textbf{e)} Example generation periods with two example double label biasing chimera states (i.e. $\{3,6\}$ (columns 1 to 3) and $\{6,9\}$ (columns 4 to 6)). The average digit state times  are shown in the bar plot on the right. 
     }\label{risultati2}
\end{figure}

We implemented two methods to obtain such chimera states, both based on the observation that the distributions of the activations of the hidden states produced through label biasing are right-skewed, with a long tail of outliers at the upper end of the distribution (see Fig.~\ref{risultati2}a). This suggests that for each state there are only a few active hidden units. 

In the first method (\textit{intersection method}), chimera states between two digits were computed by activating (i.e. $h=1$) only the units in common between the highest $k$ active units of the label biasing vectors of the two digits, while the others were set to 0. Given that we observed that the percentage of active hidden units remained constrained in a small range during the generation process (Fig.~\ref{risultati1}c), we decided to set $k$ equal to the rounded down average number of active hidden units in the first step of generation (i.e. 149). 
In the second method (\textit{double label biasing}), instead of using a one-hot encoded label for label biasing (see \cite{Zorzi et al. 2013} for details), we utilized a binary vector with two active entries (i.e. $= 1$) that corresponded to the digits of the desired chimera state. The resulting $H_{\mathrm{Label\ biasing}}$ was then binarized, keeping active only the top $k$ most active units (also here $k = 149$). Generativity (quantified as the average number of states visited in a generation period) was characterized in all intersections of two digits (100 samples per digit combination, Fig.~\ref{risultati2}b,d). Examples of chimera state generations are shown in Fig.~\ref{risultati2}c,e.

Interestingly, both techniques induced higher state exploration than the classic label biasing generation method (average number of visited states between chimera states - intersection method: $2.951 \pm 0.099$, average number of visited states between chimera states - double label biasing: $2.104 \pm 0.122$; $n = 45$ combinations of two digits), although only the intersection method state exploration was significantly higher than the classical label biasing (Mann-Whitney U test (one-sided): $p = 6.703 \cdot 10^{-5}$ (intersection method), $p = 0.139$ (double label biasing)). Some combinations of digit states (e.g. $\{6,9\}$) seemed to induce particularly high state exploration with both methods; however, the correlation between the number of visited states in the two methods was mild ($\rho = 0.334$, n=45 combinations of two digits). 
Both methods also induced a significant drop in non-digit state time (average non-digit state time - intersection method: $12.156 \pm 2.272$, Mann-Whitney U test (one-sided): $p = 2.338 \cdot 10^{-5}$; average non-digit state time - double label biasing: $25.032 \pm 4.215$, Mann-Whitney U test (one-sided): $p = 5.054 \cdot 10^{-4}$; n = 45 combinations of two digits), suggesting that the increase in exploration leads to visiting more plausible sensory states.

\section{Discussion}

In this work we introduced an original framework to study the generation dynamics of restricted Boltzmann machines, a class of generative neural networks that have been largely employed as models of cortical computation. The proposed method exploits label biasing \cite{Zorzi et al. 2013} to iteratively generate plausible configuration of hidden and visible states, thus allowing to explore the attractor landscape of the energy function underlying the generative model.

To demonstrate the effectiveness of our approach, we characterized the generation dynamics of an RBM trained on a classical dataset of handwritten digits, exploring different sampling strategies to maximize state exploration. The standard label biasing approach initiate the generation of class prototypes from the hidden representation of single digits; our simulations show that this strategy can produce high-quality digit images, but does not allow to explore multiple states during the generative process. We thus explored the possibility of initiating the generation from chimera states, which might be considered as ``meta-stable'' states that allow to reach different attractors. Both methods developed (intersection method and double label biasing) indeed increased the number of states visited during the generation process, also significantly diminishing the non-digit state time.
Nevertheless, the estimated transition matrices indicated that the non-digit state generally acts as a strong attractor, from which the system is unable to escape. This suggest that the generative dynamics of RBMs might not fully mimick the spontaneous dynamics observed in biological brains, which appear more flexible and heterogeneous.

Future work should explore more recent version of RBMs, for example the Gaussian-Bernoulli RBM \cite{Liao et al. 2022}, which is capable of generating meaningful samples even from pure noise and might thus develop more interesting generation dynamics. Another interesting research direction could be to explore more complex datasets, perhaps involving natural images, which would increase model realism and might allow to more systematically test neuroscientific hypotheses \cite{Pezzulo et al. 2021}.

%
%
%
%

\end{document}